\documentclass[runningheads]{llncs}

\usepackage[misc]{ifsym}

\usepackage{mwe}

\usepackage[utf8]{inputenc} 
\usepackage[T1]{fontenc}    
\usepackage{hyperref}       
\usepackage{url}            
\usepackage{booktabs}       
\usepackage{amsfonts}       
\usepackage{nicefrac}       
\usepackage{microtype}      
\usepackage{xcolor}         

\usepackage{amsmath}
\usepackage{amssymb}
\usepackage{mathtools}
\usepackage{style/algorithmic}
\usepackage{array}
\usepackage{tabularx}
\usepackage{fixltx2e}
\usepackage{comment}
\usepackage[caption=false]{subfig}
\usepackage{graphicx}
\usepackage{epstopdf}

\begin{document}
\title{SMOTE-DP: Improving Privacy-Utility Tradeoff with Synthetic Data}
\titlerunning{Improving Privacy-Utility Tradeoff}

\author{Yan Zhou\inst{1} \and Bradley Malin\inst{2}
 \and Murat  Kantarcioglu\inst{3}}

\institute{University of Texas at Dallas\\ Richardson, TX 75082\\
\email{yan.zhou2@utdallas.edu}
\and
Vanderbilt University Medical Center\\
Nashville, TN 37232\\
\email{b.malin@vumc.org}
\and 
Virginia Tech\\Blacksburg, VA 24061\\
\email{muratk@vt.edu}
}
\maketitle

\begin{abstract}
Privacy-preserving data publication, including synthetic data sharing, often experiences trade-offs between privacy and utility. Synthetic data is generally more effective than data anonymization in balancing this trade-off, however, not without its own challenges. Synthetic data produced by generative models trained on source data may inadvertently reveal information about outliers. Techniques specifically designed for preserving privacy, such as introducing noise to satisfy differential privacy, often incur unpredictable and significant losses in utility. In this work we show that, with the right mechanism of synthetic data generation, we can achieve strong privacy protection without significant utility loss. Synthetic data generators producing contracting data patterns, such as Synthetic Minority Over-sampling Technique (SMOTE), can enhance a differentially private data generator, leveraging the strengths of both. We prove in theory and through empirical demonstration that this SMOTE-DP technique can produce synthetic data that not only ensures robust privacy protection but maintains utility in downstream learning tasks.
\end{abstract}

\section{Introduction}

Privacy-preserving data publishing has important practical implications for decision-making in both academic and commercial applications. The main challenge of data sharing is minimizing the risk of privacy leakage so that the benefit of publicly sharing data valuable for the society will outweigh the cost of individual privacy.  Traditional methods for injecting privacy into publicly available data are based on the principle of anonymization~\cite{10.1142/S0218488502001648,10.1145/1217299.1217302}. Yet these types of techniques have been shown to be vulnerable to privacy attacks such as linkage based on residual quasi-identifiers to resources that contain the identity of the corresponding records~\cite{4531148}. More recently,  privacy-preserving data publishing methods have focused on synthetic data generation. Synthetic data has the potential to preserve statistical correlations while hiding personal  information~\cite{10.5555/3454287.3454946,10.1007/978-3-540-87471-3_20}. 

However, a recent study challenges such claims about the privacy protective nature of synthetic data~\cite{277172}. The main disagreement lies in the perception of synthetically generated data as artificial data, and whether real data can be recovered. Although synthetic data from generative models largely preserves the statistical properties of the original data, it is susceptible to the risk of linkability which allows the attacker to re-identify records in the sensitive dataset.  Individuals most vulnerable to linkage attacks are those present in the outlier group, statistically lying an abnormal distance from other samples in the population~\cite{277172}.  Using this statistical infrequency as prior knowledge, the attacker can strategically determine the membership information of a given target record in a sensitive dataset. In both image and tabular data domains, synthetic data has demonstrated certain vulnerabilities to private information extraction attacks~\cite{10.1145/3372297.3417238}. Quantifying the protection synthetic data can provide remained largely unexplored until differential privacy (DP) emerged. Differential privacy bounds the maximum impact a single data entry can have on the output of a computation. Although  differentially private models can provide formal guarantees for privacy, it is shown that they tend to incur significant utility loss~\cite{10.1007/978-3-030-57521-2_2}, and it remains unclear how the trade-off  affects the overall statistical data properties~\cite{277172}. 

This begs the question of if the two types of synthetic data approaches can be used in conjunction to complement each other, so that privacy protection and utility preservation can be achieved simultaneously. In this paper, we present a synthetic data generator that can substantially improve privacy-utility trade-off, referred to as {\em SMOTE-DP}. The SMOTE-DP generator is composed of an end-to-end construct that enables data flow from a non-DP data generator $G_1$ to a DP generator $G_2$ to produce the final synthetic dataset (as illustrated in Figure~\ref{fig:duplex}):
\begin{figure}[!htb]
\centering
\includegraphics[width=0.4\textwidth]{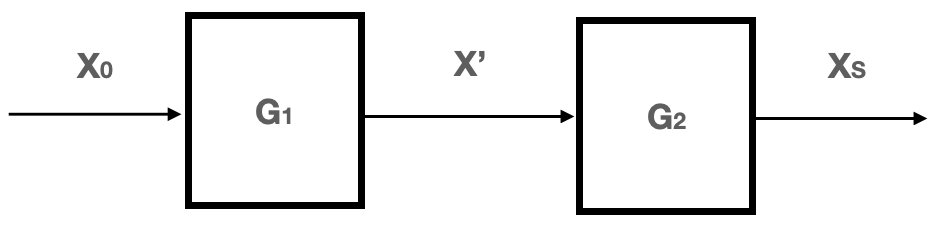}
\caption{\label{fig:duplex} SMOTE-DP: a synthetic data generator pipeline.}
\end{figure}

\begin{itemize}
\itemsep=0pt
\item[a.)] For the non-DP generator $G_1$, we employ a well-known synthetic data generator SMOTE~\cite{10.5555/1622407.1622416}. SMOTE is frequently relied upon to rebalance an imbalanced dataset to improve model accuracy on minority classes. When compared to other generative models, SMOTE has been shown to achieve a more reasonable balance between privacy and utility~\cite{NISTsynthdatatest}. To our knowledge, its power to reinforce an existing differentially private mechanism for better risk-utility trade-off has neither been studied theoretically nor empirically. Note that our goal is not to promote the use of SMOTE. As we will explain later, SMOTE can be replaced by any data generation/transformation techniques that produce contracting data patterns. 
\item[b.)] For the DP generator, we use a differentially private Bayesian network as the differential private model satisfying $\epsilon$-differential privacy~\cite{10.1145/3134428}. An excessively large privacy budget $\epsilon$, which induces very little utility loss, is often insufficient for achieving a desired privacy guarantee ~\cite{277172,10.1145/3626494}.  This scenario can be resolved by SMOTE because of the contracting nature of SMOTE-generated data patterns, which constitutes the underlying principle of SMOTE-DP  for privacy preserving synthetic data generation. As with SMOTE, any differentially private data generator can be used as a substitute for the DP Bayesian network used in this study. There is no theoretical dependence  between specific DP and non-DP generator pairs. 
\end{itemize}
An important note is that {\em SMOTE-DP is different from sequential composition where source data is consulted more than once and multiple results of differential privacy are released. SMOTE transforms  source data prior to DP.} 

We demonstrate, both in theory and through empirical analysis, that SMOTE-DP provides a more balanced risk-utility trade-off than either technique can achieve in isolation.  The main contributions of this work are as follows:
\begin{itemize}
\itemsep0em
\item We propose a synthetic data generation mechanism, composed of both DP and non-DP generators, that can radically improve utility while providing strong privacy protection.
\item We provide a theoretical justification that SMOTE can reduce sensitivity, that is, the maximum change in the output caused by a single change in the input, consequently strengthen privacy protection provided by the DP generator. 
\item We provide empirical results to demonstrate that our SMOTE-DP data generator provides strong privacy protection without sacrificing utility. 
\item We recommend to be  cautiously skeptical, because of the complexity of the real world, that higher privacy levels must be achieved with smaller privacy parameter values, and hence at the expense of significant utility loss. At least, this assumption should be verified  before it is taken for granted.   
\end{itemize}

The remainder of the paper is organized as follows. Section~\ref{sec:rl} puts our investigation in context of the related literature. Section~\ref{sec:method} presents the theory that supports our proposed technique, and   Section~\ref{sec:test} presents our empirical results. Finally, Section~\ref{sec:con} concludes our work.

\section{Related Work}
\label{sec:rl}
Data anonymization techniques such as generalization and deletion~\cite{10.1142/S0218488502001648,10.1145/1217299.1217302,4531148} were developed in response to the need of hiding personally identifiable information and hence preventing privacy leakage. These anonymization techniques were quickly shown to be vulnerable to linkage (also known as re-identification) and inference attacks~\cite{bc59d0a6d00642ae83ca6e21232a7fbd,5207644,DBLP:conf/ndss/PyrgelisTC18,7958568,10.1145/3372297.3417238,47e5cb9342984646b1167dc1aa87ac07,10.3233/JCS-191362}. 
To overcome the limitations of traditional anonymization methods, synthetic data generation techniques have been proposed~\cite{10.5555/3454287.3454946,10.1007/978-3-540-87471-3_20}. The notion of fully synthetic data was proposed~\cite{rubin1993statistical} to facilitate the release of data of ``no actual individual''. Exactly how much protection synthetic data can provide remained unanswered in techniques  covered under the conventional statistical disclosure limitation framework. And, over time, nearly all such techniques, including  ``suppression, coarsening, swapping, shuffling, sampling, and most noise-infusion techniques'', were shown to have failed to satisfy differential privacy~\cite{10.1007/978-3-540-87471-3_20}, especially when data is high-dimensional and sparse and the validity of protection can only be guaranteed when the minimum prior sample size is much larger than practically possible~\cite{GehrkeKiferMachanavajjhalaAbowdVilhuber08_PrivacyTheoryMeetsPracticeOnMap}. 

Differential privacy~\cite{10.1007/11787006_1,Hall_Wasserman_Rinaldo_2013,8894030,mironov2017renyi,10.1145/3110254} (DP), compared to traditional statistical closure limitation, provides theoretical guarantees against inferential disclosure for synthetic data. DP bounds the maximum amount of information an attacker can infer from the presence of an individual.  
Despite its great popularity, DP's risk-utility trade-off, where privacy risk and data utility are inversely correlated, has led to some concerns~\cite{10.1007/978-981-19-7769-5_13,10.1007/978-3-030-57521-2_2,277172}. Recently, Muralidhar et al.~\cite{10.1007/978-3-030-57521-2_2} raised a strong objection to the use of DP as a privacy model on microdata by demonstrating that it does not guarantee both confidentiality and utility, and the amount of information loss as a result of privacy protection remains unknown. DP synthetic data has also been shown to be vulnerable to privacy attacks such as linkage attacks~\cite{DBLP:journals/corr/abs-2011-07018}. More important, it has been observed that some individuals receive substantially less protection than others and, perhaps of greater concern, it is difficult to predict which individuals records are vulnerable to attack. Stadler et al.~\cite{277172} restated similar concerns and proposed new privacy protection measures. They point out that differentially private synthetic data protects individuals at a significant cost in utility, and it lacks transparency about the trade-off. As a result, it remains unclear what data properties have been preserved and what have been suppressed. 
The risk-utility trade-off also creates a debate about what the ideal privacy parameter $\epsilon$ should be. NIST's 2022 challenge on the selection of $\epsilon$ reported that organizations choose a wide range of $\epsilon$ values, all significantly exceeding the value recommended by the DP researchers~\cite{nistblog20220124}.

Regardless of the existing open challenges, DP models have been quickly adopted by an increasing number of organizations in industry~\cite{desfontainesblog20211001}. Addressing the risk-utility issue remains a highly non-trivial matter. In this paper, we present a synthetic data generation mechanism that radically improves the privacy-utility tradeoff. The proposed technique is composed of a differential private data generator and the oversampling technique  SMOTE~\cite{10.5555/1622407.1622416}. The idea is motivated by the recent findings that  new samples synthesized with SMOTE demonstrate better utility and good privacy compared to other synthetic data generators~\cite{NISTsynthdatatest}. In addition, SMOTE-based oversampling has also been demonstrated as an effective technique for mitigating biases in AI algorithms~\cite{doi:10.1137/1.9781611977653.ch98,10.1145/3468264.3468537,gonzlez_zelaya_parametrised_2019,10.1145/3340531.3411980,lavalle2022methodology}. On the one hand, SMOTE demonstrates characteristics in line with differential privacy models that are protective of the private/sensitive data to an extent; on the other hand, SMOTE has the potential to complement differential privacy by keeping utility loss bounded. 

SMOTE is widely used in machine learning to oversample a minority class to improve classification performance when data is imbalanced.  SMOTE first selects an instance $x$ from the minority class, then creates synthetic samples from $x$ by performing linear interpolation between $x$ and one of its $K$ nearest neighbors $x_i \in {x_1, \ldots, x_K}$. Informally, SMOTE generates data in accordance with the same assumption that the $k$-nearest neighbor algorithm is based on. In this respect, similar points are located near one another and share the same density function value. One disadvantage of SMOTE-generated synthetic data is a potential to deviate from the true distribution of the underlying group~\cite{10.1007/978-3-540-28645-5_30}. SMOTE-generated samples tend to be more contracted and form unrealistic graph patterns  where edges are densely populated with synthetic samples due to linear interpolation~\cite{10.1007/s10994-022-06296-4,ELREEDY201932}.  However, SMOTE and its extensions in general produce better classification accuracy. This benefit is not limited to minority data. In fact, both accuracy and distribution deviation improve when the size of data increases~\cite{ELREEDY201932}. 

SMOTE with differential privacy was empirically confirmed for its positive effect on downstream classification where a privately generated noisy histogram was used~\cite{dpsmotethesis}. To develop a deeper understanding of the effect of SMOTE on a DP generator,  in this paper we directly tackle the underlying mechanism that connects to the improved risk-utility trade-off in the context of privacy attacks such as linkage attacks. 

\section{Methodology}
\label{sec:method}

In this section, we provide theoretical justification for our data generation mechanism, SMOTE-DP, that improves privacy and preserves utility. In essence, SMOTE generates contracting data patterns with smaller covariances that allow differentially private mechanisms to choose ``unreasonably" large $\epsilon$ values that would normally be considered as meaningless. Large $\epsilon$ values typically correspond to an underestimation in sensitivity,  raising the potential for breaking differential privacy guarantees. We explain how SMOTE assists in mitigating this problem. 

\subsection{Background}
Our proof is built on two important foundations. One is the sensitivity of query output.  The other is the distribution of SMOTE-generated data, which verifies contracting data patterns generated by SMOTE. 

\subsubsection{Sensitivity of Query Output}

A differentially  private algorithm achieves its bound by applying an amount of noise that scales with the sensitivity of the query output. Given a query function $f$, sensitivity measures the maximum change in the output of $f$ caused by a single change to an instance in the dataset. Formally, sensitivity is defined as follows:
\begin{definition}
Given a query function $f: D^n \rightarrow R$, let the sensitivity of $f$ be $\Delta f$:
\[
\Delta f = \sup_{X,X':d(X,X') \le 1} | f(X) - f(X')|
\]
\end{definition}
where $d(X,X')$ is the distance between two datasets $X$ and $X'$, and two neighboring datasets have a distance no greater than one.

\subsubsection{SMOTE-generated Data Patterns} 
\label{sec:smotecontraction}

\begin{figure}[!ht]
\centering
\subfloat[]{\includegraphics[width=.3\textwidth]{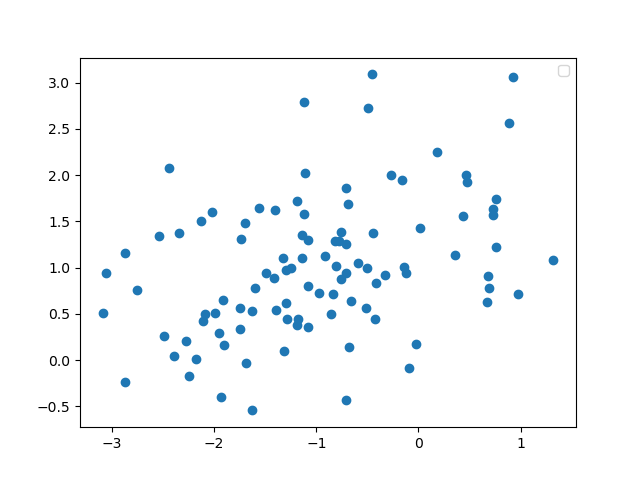}}\quad
\subfloat[]{\includegraphics[width=.3\textwidth]{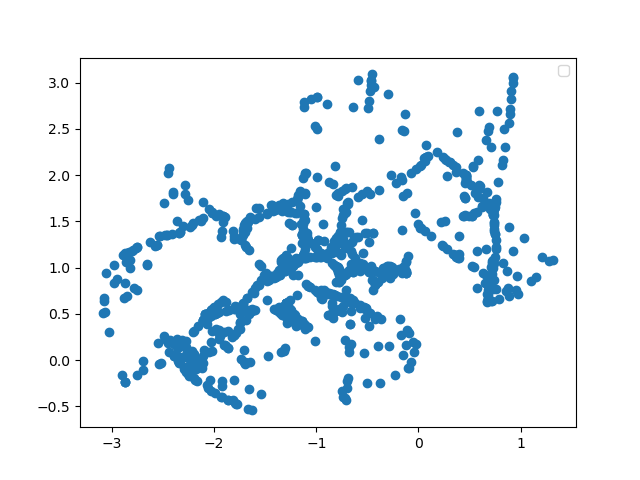}}
\caption{\label{fig:contraction} Original data distribution vs. SMOTE-generated sample distribution.}
\end{figure}

Theoretical analyses have been developed to help understand the underlying workings of SMOTE.  Elreedy and Atiya~\cite{ELREEDY201932} explore SMOTE-generated data patterns by investigating the mean and covariance of the new distribution. Let $x_0 \in R^d$ be a data point in the original distribution, and $x$ be one of its $K$ nearest neighbors. Let their difference $\Delta = x - x_0$.  The SMOTE-generated sample $z$ is $z = x_0 + w\Delta$, where $w$ is a random number from the uniform distribution in $[0,1]$.  Assuming the probability density of the original distribution is multivariate Gaussian, the mean and the covariance of the SMOTE-generated data distribution are~\cite{ELREEDY201932}:
\begin{flalign}
\label{eq:dsmote}
E(z) & \approx  \mu_{x_0}  & \\
\Sigma_z & =   \Sigma_{x_0}  +  [(2\pi)^{\frac{1-d}{2}} \frac{C}{3} |\Sigma_{x_0}|^{\frac{1-d}{2d}} \Sigma_{x_0} (\frac{d}{2d-1})^{\frac{d}{2}} - 2\pi C |\Sigma_{x_0}|^{\frac{1}{d}}\Sigma_{x_0}(\frac{d}{d-2})^{\frac{d+2}{2}}] I 
\end{flalign}
where $C$ is a random variable parameterized in terms of the dimension $d$ and the number of chosen neighbors $K$, $|\Sigma_{x_0}|$ is the determinant, and $I$ is the identity matrix. Note that the mean of the SMOTE-generated samples is close to the true mean. Algebraic computation results in a negative value from the term inside the square brackets in the covariance, suggesting the contraction of the SMOTE-generated data patterns compared to the original distribution, with the diagonal variances becoming smaller, as illustrated in Figure~\ref{fig:contraction}. Hence,
\begin{equation}
\label{eq:covsmote}
tr(\Sigma_z) < tr(\Sigma_{x_0})
\end{equation}
Other probability density assumptions such as multivariate Laplace also suggest such contracting effect on SMOTE-generated data. Later, Elreedy et al. formulate the probability distribution of the SMOTE-generated samples~\cite{10.1007/s10994-022-06296-4},  
which is also in consistent with the contracting phenomenon derived in~\cite{ELREEDY201932}. 

\subsection{SMOTE-DP: Theory and Proof}
Given a dataset $X \sim \mathcal{N}(\mu,\Sigma)$, let $X' \sim \mathcal{N}(\mu,\Sigma)$ be a dataset such that $X$ and $X'$ differ by one instance, denoted as $||X-X'||_1 \le 1$. We define the sensitivity $\Delta f_s$ of the synthetic output as follows:
\begin{definition}
\label{def:sen}
For any datasets $X,X': ||X-X'||_1 \le 1$, a generative model $f_S$, and two synthetic data sets, $X_S=f_S(X)$ and $X'_S=f_S(X')$, generated from $X$ and $X'$ with $f_S$, the sensitivity of the synthetic output is:
\[
\Delta f_s = \max_{X_S, X'_S}\mathop{\mathbb{E}}(||X_S-X_S'||) = \max_{X_S, X'_S}(||\mu_{X_S} - \mu_{X'_S}|| + tr(\Sigma_{X_S} + \Sigma_{X_S'}))
\]
where  $X_S, X'_S \sim \mathcal{N}(\mu_S,\,\Sigma_S)$, $\mu_{X_S}$ and $\mu_{X_S'}$ are the mean of $X_S$ and $X'_S$, and $\Sigma_{X_S}$ and  $\Sigma_{X'_S}$ are the covariance matrices. 
\end{definition}

\subsubsection{Sensitivity of SMOTE-transformed Data}
\label{sec:sensmote}
Given a dataset $X$, we apply SMOTE on $X$ to generate synthetic data sets $X_S$ and $X_S'$ where $||X_S-X_S'||_1 \le 1$. A DP algorithm achieves $\epsilon$-differential privacy by adding noise to $X_S$ according to a Laplace distribution:
\begin{equation}
\label{eq:dp}
X^{\epsilon} = f_{S}(X) + Y, \text{where}\; Y \sim Lap(\Delta f_s /\epsilon)
\end{equation}
where $\Delta f_s$ is the sensitivity defined on the synthetic data generation mechanism $f_S()$. In this paper, $f_S()$ is specifically defined as SMOTE. Notice the amount of noise introduced scales proportionately to  sensitivity. 
\begin{theorem}
\label{thm:sensitivity}
(SMOTE Sensitivity Reduction)
Let $\Delta f$ be the sensitivity defined on dataset $X$, $\Delta f_s$ be the sensitivity on the SMOTE-transformed dataset $X_S$ with $f_{S} : X \mapsto X_{S}$, $\Delta f_S < \Delta f$ given $X \sim \mathcal{N}(\mu,\,\Sigma)$ and  $X_S \sim \mathcal{N}(\mu_S,\,\Sigma_S)$.
\end{theorem}
\begin{proof}
Given any datasets $X,X': ||X-X'||_1 \le 1$, according to Definition~\ref{def:sen}, its sensitivity is:  
\[
\Delta f = \max_{X,X'}\mathop{\mathbb{E}}(||X-X'||) = \max_{X,X'}(||\mu_{X} - \mu_{X'}|| + tr(\Sigma_{X} + \Sigma_{X'}))
\]
where $\mu_{X},\,\Sigma_{X}$ are the mean and covariance of $X$ and   $\mu_{X'},\,\Sigma_{X'}$ are the mean and covariance of $X'$. 
 Hence, 
 \[
 \Delta f_{s} - \Delta f =  \max_{X_S,X'_S}( ||\mu_{X_{S}} - \mu_{X'_{S}}|| + tr(\Sigma_{X_{S}} + \Sigma_{X_{S}'})) - \max_{X,X'}( ||\mu_{X} - \mu_{X'}|| + tr(\Sigma_{X} + \Sigma_{X'}))\]
 According to Eq. (1) and the inequality~(\ref{eq:covsmote}) in Section~\ref{sec:smotecontraction}, 
\[ \max_{X_S,X'_S} ||\mu_{X_{S}} - \mu_{X'_{S}}|| - \max_{X,X'}||\mu_{X} - \mu_{X'}|| \approx 0, \]
\[\max_{X_S,X'_S}tr(\Sigma_{X_{S}} + \Sigma_{X_{S}'}) - \max_{X,X'}tr(\Sigma_{X} + \Sigma_{X'}) < 0. \]
 Therefore, $\Delta f_{s}  < \Delta f$, that is, the sensitivity is reduced after applying SMOTE.
 \end{proof}

We now prove SMOTE can enhance privacy by essentially reducing the sensitivity of the original dataset $\Delta f$, that is, $\Delta f_{s} = \alpha \cdot \Delta f$ where $0< \alpha <1$ because of SMOTE's ability to produce contracting data patterns.

\subsubsection{Privacy of SMOTE-DP}
With the definition of sensitivity for a synthetic generator, we prove a theorem that states SMOTE can enhance privacy given a privacy budget $\epsilon$ in DP post-processing. 
\begin{theorem}
\label{thm:dp-smote}
 (SMOTE Enhanced Privacy)
 Given a SMOTE-transformed dataset $X_S$, $X_{S}^{\epsilon}$---   generated with an $\epsilon$-differential private mechanism---is differentially private with respect to $\alpha \cdot \epsilon$ in the $\Sigma_{X_{S}}$-transformed metric space, where $0 < \alpha < 1$, and $\Sigma_{X_{S}}$ is the covariance matrix of SMOTE-generated data. 
\end{theorem}
\begin{proof}
From the discussion in Section~\ref{sec:sensmote}, we know SMOTE-generated patterns are contracting with diagonal variances $(\sigma_1, \ldots, \sigma_n)$ becoming smaller in $\Sigma_{X_{S}}$ than in $\Sigma_{X}$. Hence, with SMOTE preprocessing, the amount of noise for achieving $\epsilon$ differential privacy should be calibrated with respect to the updated {\em sensitivity} $\Delta f_{s}$.  Since  $\Delta f_{s}  < \Delta f$, or equivalently, $\Delta f_{s} = \alpha \cdot \Delta f$ where $0< \alpha <1$, it follows: 
\[\frac{\Delta f}{\epsilon} = \frac{\Delta f_{s}}{\alpha \cdot \epsilon} \]

Mapping the $\epsilon$-differentially private data $X^{\epsilon}$ to the SMOTE-transformed data space $X^{\epsilon}_S$, the noise introduced to the original data $X$ to achieve $\epsilon$-differential privacy effectively becomes the noise added to $X$ to achieve $(\alpha \cdot \epsilon)$-differential privacy on $X_{S}$:
\begin{equation}
\label{eq:dp-sm}
X_{S}^{\epsilon} = f_{S}(X) + Y + w\Delta, \text{where}\; Y \sim Lap({\Delta f_{s}} /(\alpha \cdot \epsilon)) 
\end{equation}
which essentially decreases the privacy budget from $\epsilon$ to $\alpha \cdot \epsilon$ for $0 <  \alpha < 1$. 
\end{proof}

Theorem~\ref{thm:dp-smote} justifies that an unreasonably large $\epsilon$ chosen as a compromise for better utility for generating $X^{\epsilon}$ (Eq.~\ref{eq:dp}) can  become a reasonable choice on the SMOTE preprocessed substitute $X_{S}$ (Eq.~\ref{eq:dp-sm}), demonstrating the benefit of post-processing DP on the SMOTE-transformed data in synthetic data publishing. 

\section{Experimental Results}
\label{sec:test}

All experiments presented in this section were conducted on an Intel\textsuperscript{\textregistered} Xeon\textsuperscript{\textregistered} machine with a 2.30GHz CPU, 256GB memory,  and four 8GB GeForce RTX 2080 GPUs. 
We first verify SMOTE can enhance DP with better utility on a simple artificial dataset for which we know its population mean and covariance. 
\subsection{Experiments on Artificial Data}
\begin{figure}[!htb]
\centering
\begin{minipage}{0.245\textwidth}
\centering{Original}
\includegraphics[width=\textwidth]{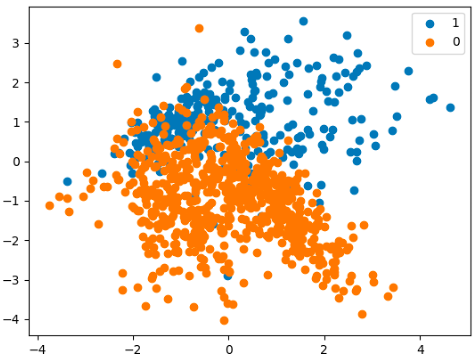}
\centering{\scriptsize{Mean\\}}
\centering{\scriptsize{Cov}}
\end{minipage}
\begin{minipage}{0.245\textwidth}
\centering{DP}
\includegraphics[width=\textwidth]{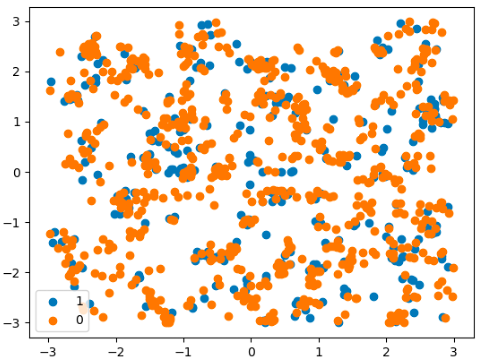}
\centering{\scriptsize{Mean F-norm: 0.375}}
\centering{\scriptsize{Cov F-norm: 4.154}}
\end{minipage}
\begin{minipage}{0.245\textwidth}
\centering{SMOTE}
\includegraphics[width=\textwidth]{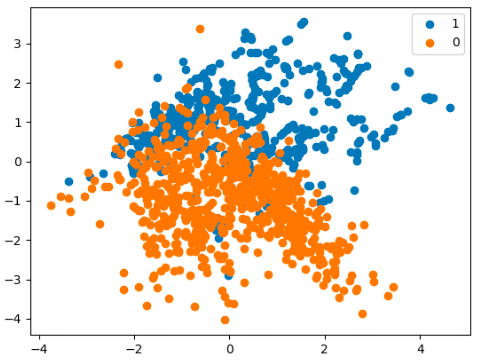}
\centering{\scriptsize{Mean F-norm: 0.399}}
\centering{\scriptsize{Cov F-norm: 0.291}}
\end{minipage}
\begin{minipage}{0.245\textwidth}
\centering{SMOTE-DP}
\includegraphics[width=\textwidth]{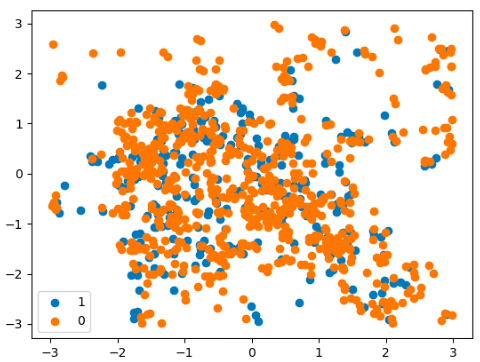}
\centering{\scriptsize{Mean F-norm: 0.356}}
\centering{\scriptsize{Cov F-norm: 2.727}}
\end{minipage}
\caption{\label{fig:artificial} Data distributions before and after synthetic generation with DP, SMOTE, and SMOTE-DP.}
\end{figure}

We generate a simple two-dimensional dataset $X$ with two classes, each following a normal distribution.  We investigate whether applying a DP mechanism on  SMOTE-preprocessed data $X_S$ produces better estimate of the mean $\mu$ and covariance $\Sigma$ than applying DP on the original non-private $X$. We measure the difference in mean and covariance using Frobenius norm. Figure~\ref{fig:artificial} shows the data distributions before and after synthetic generation with DP, SMOTE, and SMOTE-DP, and the Frobenius norm (denoted as F-norm in the figure) of the mean and covariance differences between the original and the synthetic data generated using the three mechanisms. As can be seen, DP-generated data has the highest covariance difference (4.154) from the original, SMOTE-processed data has the smallest covariance difference (0.291), while SMOTE-DP discovers the middle ground (2.727) between DP and SMOTE. All three result in approximately the same norm for the mean. The experiment demonstrates that SMOTE enhances DP with better data utility by respecting the true mean and covariance. Similar experiments were repeated on 10-dimensional artificial datasets with a various number of informative attributes. The same conclusion can be drawn.  

\subsection{Experiments on Real Data}
We now focus on test cases reported in~\cite{277172} that were claimed to be most challenging where DP has experienced significant utility loss while protecting privacy, and other generative models with no explicit privacy protection fail to protect outlier data points in the face of linkage attacks. 
We adapted their source code~\footnote{https://github.com/spring-epfl/synthetic\_data\_release} to repeat their experiment and performed additional experiments with SMOTE for comparison. 
\subsubsection{Experiment Setup} Details of datasets, generative models, and metrics used in this experiment are discussed below. 

{\bf Datasets} We include three datasets {\em Texas}~\cite{texasdataset}, {\em German Credit Risk}, and {\em Employee}~\cite{kaggledataset} 
in the experiments. 
The {\em Texas} dataset is the hospital discharge data file released by the Texas Department of State Health Services. As in~\cite{277172}, the dataset used in this experiment consists of 50,000 randomly sampled patient data. There are 18 attributes including 11 categorical and 7 continuous attributes. 
The {\em German Credit} data, for greater clarity, contains a smaller set of attributes compared to the original source in the UCI repository~\cite{misc_statlog_(german_credit_data)_144}.  It has ten attributes including the class label. The total number of entries in the dataset is 1000, and each entry represents a person labeled as good or bad credit risk. 
The {\em Employee} data contains the information of 4,653 employees in a company. There are nine attributes including the class label. 

{\bf Generative Models} 
We use the same generative models, both private and non-private, as in~\cite{277172}. GreedyBayes~\cite{10.1145/3085504.3091117} is a non-private implementation of Bayesian network. The sanitization procedure by NHS England~\cite{NHS} implemented in~\cite{277172} is used for comparing the effectiveness of synthetic data and data sanitization. Differentially  private generative models used in our experiment include PrivBayes~\cite{10.1145/3134428} and PATEGAN~\cite{yoon2018pategan}. An extension of SMOTE, SMOTENC, implemented in the Python library~\cite{SmoteNC} is used both standalone and in SMOTE-DP with parameter $K = 1$. 

{\bf Privacy Gain} Protection is measured in terms of privacy gain defined in~\cite{277172} as the reduction in advantage when the adversary is provided with synthetic data $S$ instead of the real data $X$:
$PG = A(X,x_t) - A(S, x_t)$
where $A$ estimates the adversary's advantage and 
$x_t$ is the target of protection. 

{\bf Attack Target} As in~\cite{277172}, five outlier records (outside the 95\% quantile range) that are most vulnerable to privacy attacks are chosen for our experiments. Besides outliers, random samples are also selected and studied for comparison.  

\subsubsection{Linkage (Membership Inference) Attack and Utility}
Linkage attacks are modeled as a game played between a data publisher and an adversary~\cite{8429311,DBLP:conf/ndss/PyrgelisTC18,277172}. Given publicly released data $S$ and some prior knowledge, the adversary aims to decide whether a target $x_t$ is in the source data $X$. 

In a linkage attack, if the source data $X$ is released, the advantage of the adversary is assigned the highest value of one since the adversary only needs to verify the presence of $x_t$ in the released $X$. The privacy gain is defined as $PG = 1 - A(S, x_t)$ where $A(S,x_t)$ is the adversary's advantage if $S$ is released, and is defined as the difference in the probability that the adversary's guess is ``yes'':
$\displaystyle A(S, x_t) = P(Yes | S, x_t \in X) - P(Yes | S, x_t \notin X)$.
Hence, if the privacy gain $PG \ge 1$ by releasing the synthetic data $S$, the target $x_t$ is protected; otherwise, there is privacy risk. $PG > 1$ suggests, occasionally, training a generator without the target $x_t$ actually improves the adversary's chance of predicting ``Yes''. 

\begin{figure}[!htb]
\begin{minipage}{0.475\textwidth}
\centering
\includegraphics[trim={0.1in 0 0 0.1in},clip,width=0.675\textwidth]{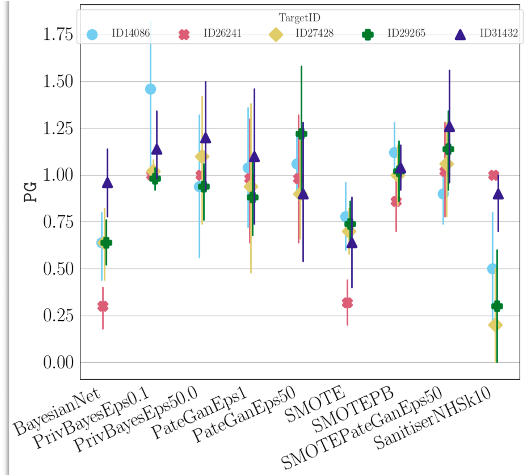}
\end{minipage}
\hfill
\begin{minipage}{0.475\textwidth}
\centering
\includegraphics[width=0.8\textwidth]{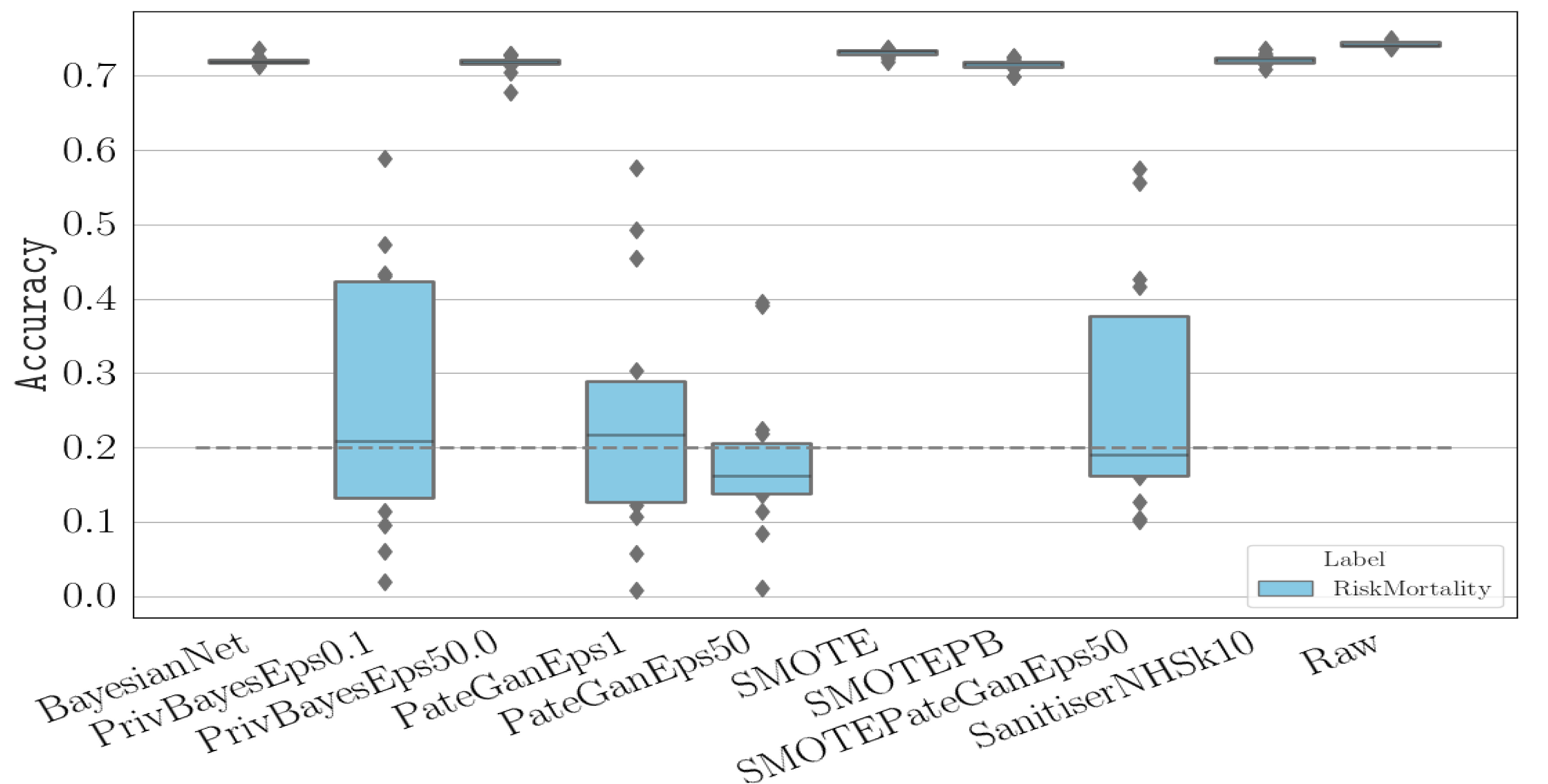}
\end{minipage}
\caption{\label{fig:texas_link_outlier} Adversary's privacy gain \& utility on five outliers in the {\em Texas} dataset.}
\end{figure}

The work reported in~\cite{277172} argued that DP algorithms failed to protect outliers and privacy was achieved at the expense of utility. Our experiment suggests more careful evaluations, or at least extensive studies are needed to warrant such a claim.   
The left plot in Figure~\ref{fig:texas_link_outlier} illustrates the privacy gain (PG) on five manually selected outliers, identical to the ones reported in~\cite{277172}, when different data release techniques are used to publish the {\em Texas} dataset. The right plot shows the \textit{utility in terms of classification accuracy with the Random Forest classifier}. The adversary is assumed to apply a feature selection technique using correlation analysis~\cite{277172}. In this experiment, besides setting $\epsilon \in\{0.1,1.0\}$ as normally being done, we added an additional study where we set an outrageously large $\epsilon=50$. Quite counterintuitively, the privacy of the selected outliers are equally well protected ($PG \approx 1$) with such a high $\epsilon$ value using the PrivBayes algorithm ($2^{nd}$ \& $3^{rd}$ from left to right in the plots). The corresponding utility, as a result of such high $\epsilon$, is nearly as good as what has been achieved on the non-private raw data. Hence, it is likely that choosing much higher $\epsilon$ values is sufficient to  both protect privacy and retain utility on this group of outliers. PateGAN also appeared to be insensitive to the $\epsilon$ value on the outliers ($4^{th}$ \& $5^{th}$ in the plots). 

When SMOTE is used to preprocess the non-private data with $\epsilon=50$, SMOTE-PateGAN ($2^{nd}$ from the right)  achieved much better privacy than  PateGAN ($\epsilon=1.0$). SMOTE helped improve the utility of PateGAN, however, the loss is still significant potentially because of mode collapse introduced by PateGAN.  SMOTE-PrivBayes (SMOTEPB, $3^{rd}$ from the right) achieved similar privacy protection as PrivBayes ($\epsilon=0.1$) except for one outlier, however, SMOTE-PrivBayes retained much higher utility than PrivBayes did. All DP generators provided better privacy than the non-DP generators ($1^{st}$, $6^{th}$ \& last).

\begin{figure}[!htb]
\begin{minipage}{0.475\textwidth}
\centering
\includegraphics[trim={0.1in 0 0 0.1in},clip,width=0.675\textwidth]{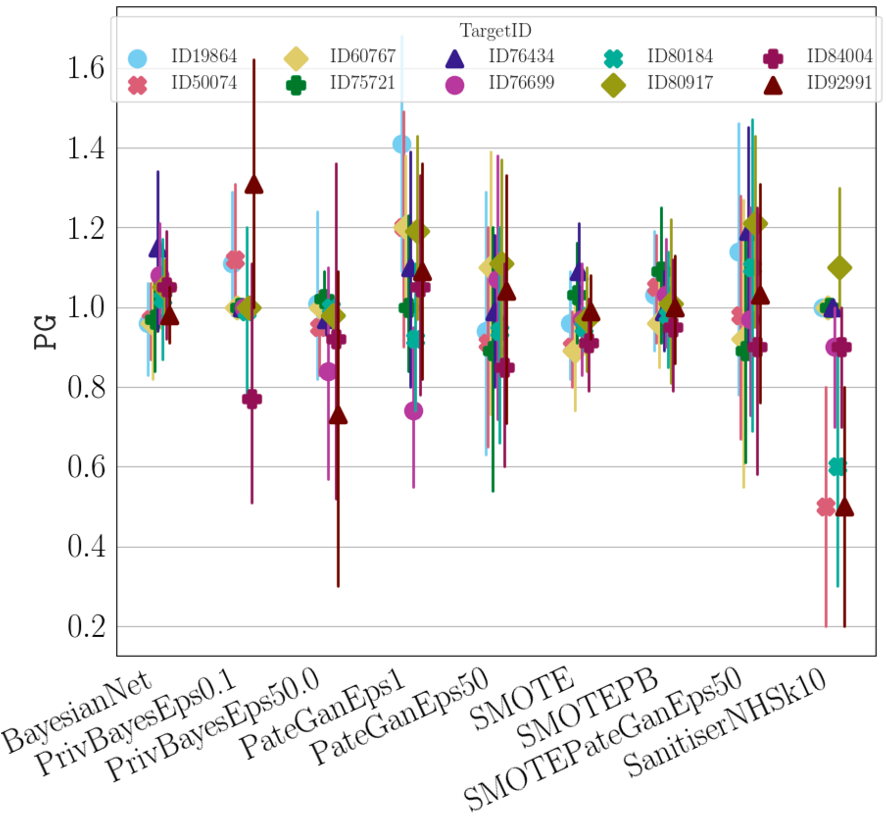}
\end{minipage}
\hfill
\begin{minipage}{0.475\textwidth}
\centering
\includegraphics[width=0.8\textwidth]{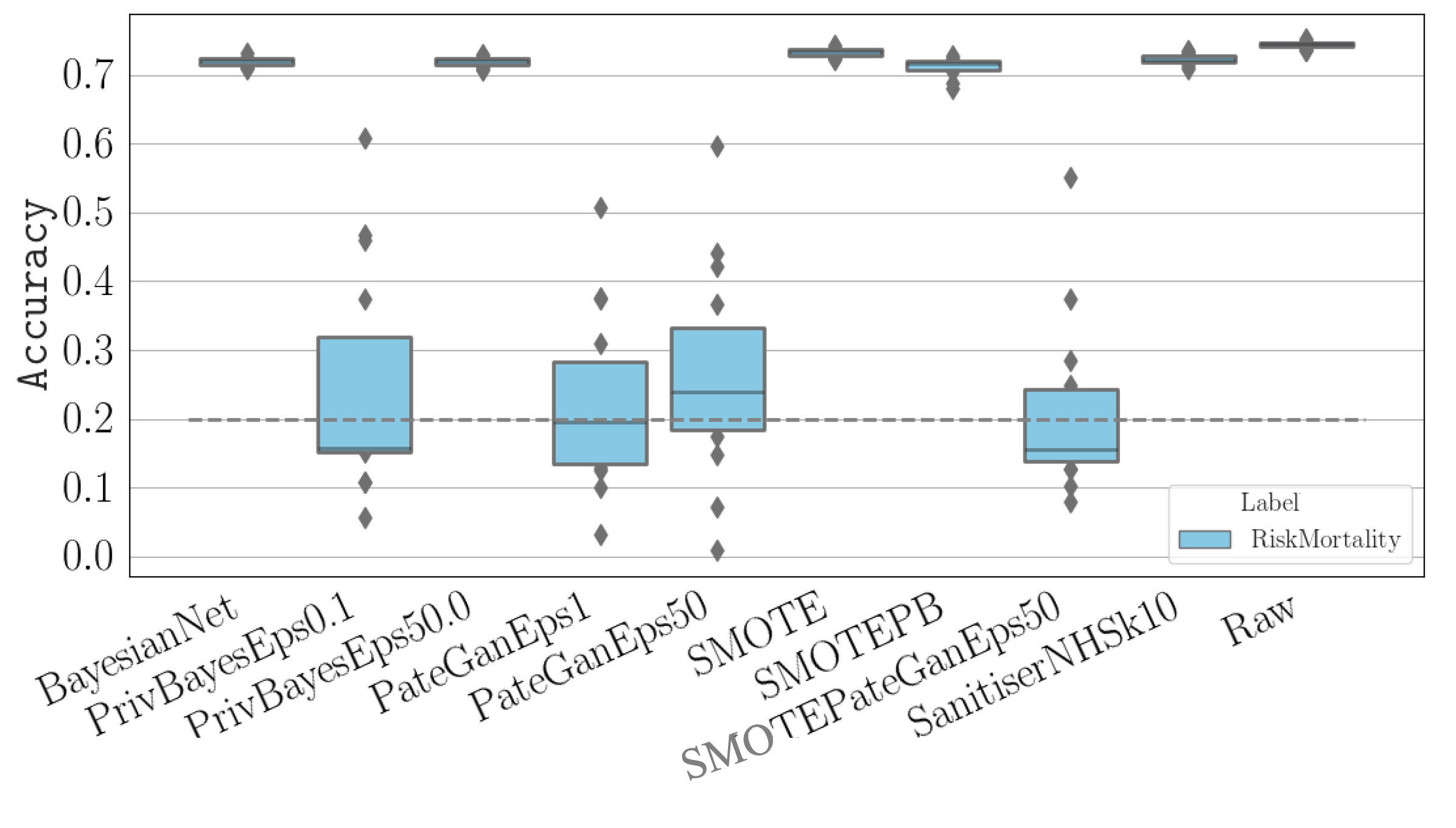}
\end{minipage}
\caption{\label{fig:texas_link_rand} Adversary's privacy gain \& utility on 10 random samples in the {\em Texas} dataset.}
\end{figure}

We also recommend to be cautious when considering DP for privacy protection. Figure~\ref{fig:texas_link_rand} shows the results of repeating the previous experiment using 10 randomly selected instances to replace the five outliers. {\em The non-DP generator BayesianNet ($1^{st}$ from the left) provided much better privacy protection than all DP generators ($2^{nd}$--$4^{th}$) without incurring any utility loss.} Note that SMOTE-preprocessing helped DP generators achieve comparable levels of privacy to BayesianNet. 

As seen in Figure~\ref{fig:texas_link_rand}, smaller $\epsilon$ values ($2^{nd}$ and $4^{th}$) provided better privacy for non-outliers than the larger $\epsilon=50$ ($3^{rd}$ and $5^{th}$) as we normally expect, at the expense of utility. When PrivBayes with $\epsilon=50$ was applied to the SMOTE-preprocessed data, better privacy ($PG \approx 1$) and stability (low variance) were observed, while utility was largely retained with very little loss. This confirms our hypothesis that SMOTE can enhance both privacy (tolerating large DP $\epsilon$ values) and utility (negligible utility loss) for the post-processing DP mechanism. 

On a side note, we do not have to designate SMOTE to transform the non-private data space. As in our proof, any synthetic data generators producing contracting data patterns, i.e., data with smaller covariance, can serve as a privacy/utility enhancement mechanism for any DP algorithms, by setting a much less aggressive privacy budget (large $\epsilon$ values).   

\begin{figure}[!htb]
\begin{minipage}{0.475\textwidth}
\centering
\includegraphics[trim={0.1in 0 0 0.1in},clip,width=0.8\textwidth]{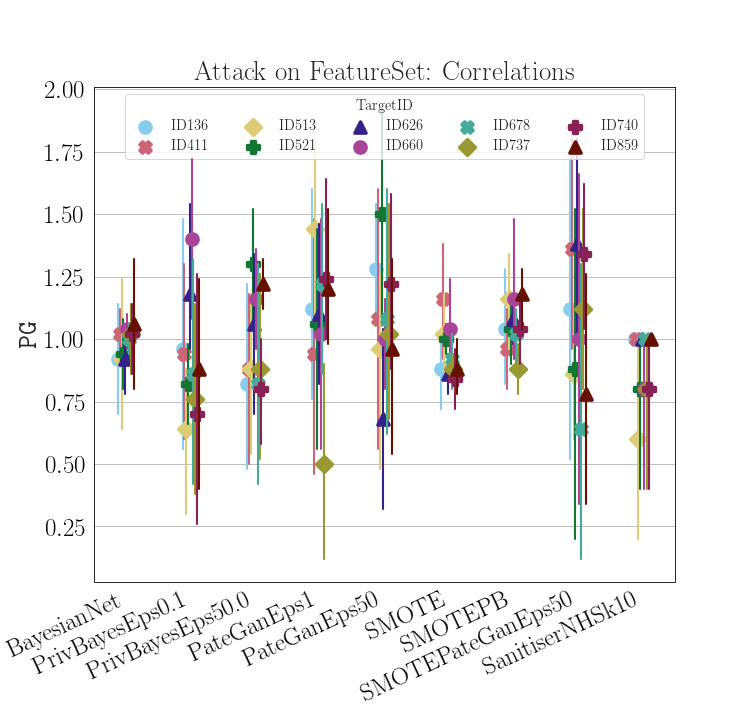}
\end{minipage}
\hfill
\begin{minipage}{0.475\textwidth}
\centering
\includegraphics[width=0.8\textwidth]{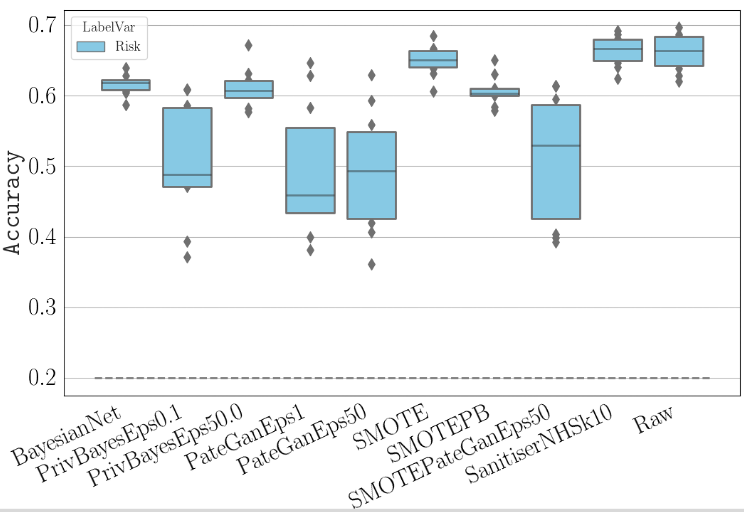}
\end{minipage}
\caption{\label{fig:german_link_rand} Adversary's privacy gain \& utility on 10 random samples in the {\em German Credit} dataset with different publishing methods.}
\end{figure}

Similar observations can be made on the $German Credit$ datset, shown in Figure~\ref{fig:german_link_rand}. When DP generators are applied to the SMOTE-preprocessed data, stable privacy protection (low variance on all samples) can be achieved without high utility loss. Pate-GAN again experienced mode collapse, making it a less ideal candidate for privacy protection on this dataset. 

\begin{figure}[!htb]
\begin{minipage}{0.475\textwidth}
\centering
\includegraphics[trim={0.1in 0 0 0.1in},clip,width=0.675\textwidth]{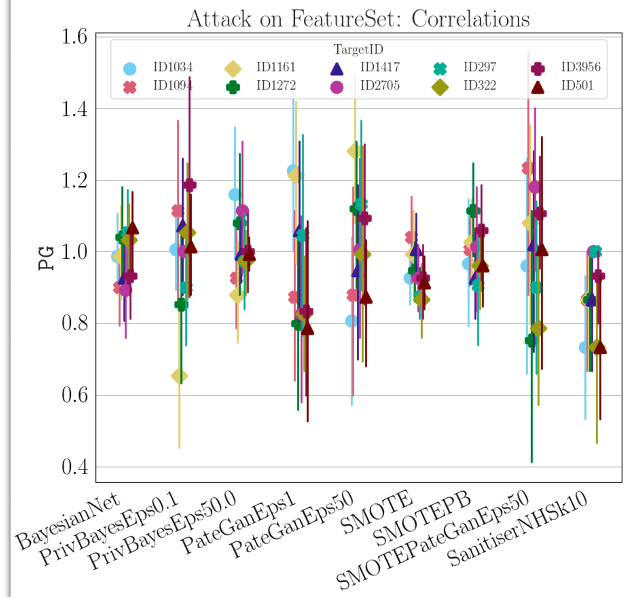}
\end{minipage}
\hfill
\begin{minipage}{0.475\textwidth}
\centering
\includegraphics[width=0.8\textwidth]{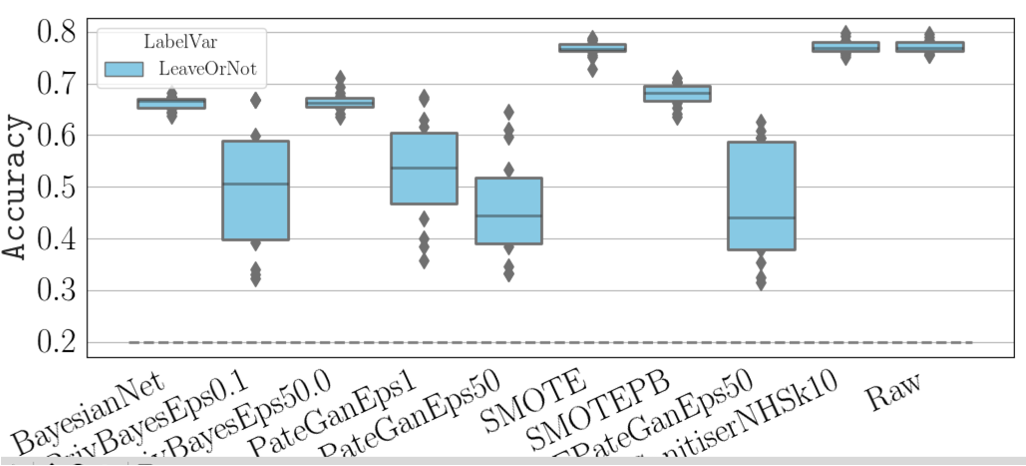}
\end{minipage}
\caption{\label{fig:employee_link_rand} Adversary's privacy gain \& utility on random samples in the {\em Employee} dataset with different publishing methods.}
\end{figure}

Results on the {\em Employee} dataset present another counterintuitive aspect of the DP algorithms on the randomly selected samples, as shown in Figure~\ref{fig:employee_link_rand}. With the large $\epsilon=50$, the DP algorithms actually achieved better privacy than with a much smaller $\epsilon \in \{0.1,1.0\}$. In addition, Non-private BayesianNet appeared to preserve privacy better than the two DP algorithms. Again, when the DP algorithms are applied to SMOTE-preprocessed data, they provide better stability and privacy-utility tradeoff.  

All our experiments suggest DP alone cannot provide consistent and stable privacy protection for all individuals in real world applications. It is not uncommon to experience worse privacy protection for some individuals with the use of DP mechanisms. However, our SMOTE-DP technique has been shown to consistently provide better and stable privacy protection without significantly hurting utility.  
\section{Conclusions}
\label{sec:con}
 The release of synthetic data has been touted as offering enhanced privacy protection compared to traditional data sanitization techniques. However, recent research has begun to question the prevailing optimism surrounding the use of synthetic data for privacy-preserving data publishing.
One issue is that synthetic data produced by non-private generative models may fail to protect the privacy of outlier records in the dataset. Conversely, while differentially private generators offer strong protection against privacy breaches, they often do so at the expense of significant utility loss. To address this issue, we propose a data generation mechanism that combines the existing non-private synthetic data generator SMOTE with differentially private generative models, aiming to improve utility without compromising privacy protection.
The idea is to set a very large privacy budget $\epsilon$ for the differentially private generator. 
With the contracting nature of data processed by SMOTE (smaller covariance), 
allocating a larger privacy budget becomes justifiable, effectively preserving both privacy and high utility of the data. Through both theoretical investigation and empirical demonstration, we show that our proposed data generation mechanism, SMOTE-DP, significantly enhances data utility while still safeguarding sensitive individual information.

\subsubsection{\discintname}
The authors have no competing interests to declare that are relevant to the content of this article.


 
\bibliographystyle{splncs04}
\bibliography{synpriv}





\end{document}